\documentclass[runningheads]{llncs}
\usepackage{amsmath}
\usepackage{graphicx}
\usepackage{caption}
\usepackage{subcaption}
\usepackage{amsfonts}
\usepackage{bm}
\usepackage{subcaption}
\usepackage{float}
\usepackage{algorithm}
\usepackage[noend]{algpseudocode}

\usepackage{hyperref}

\newcommand{\MATLAB}{\textsc{Matlab}\xspace}
\usepackage{xspace}

\DeclareMathOperator*{\argmin}{arg\,min}

\begin{document}
\title{A Multi-channel DART Algorithm}
\titlerunning{A Multi-channel DART Algorithm}

%
%
\author{Math\'e Zeegers\inst{1} \and
Felix Lucka\inst{1,2} \and
Kees Joost Batenburg \inst{1}}
\authorrunning{M. Zeegers \and F. Lucka \and K.J. Batenburg}
%

\institute{Computational Imaging, Centrum Wiskunde \& Informatica (CWI), Amsterdam, The Netherlands \and
Department of Computer Science, University College London, UK}

\maketitle              

\vspace{-1cm}
\begin{abstract}
Tomography deals with the reconstruction of objects from their projections, acquired along a range of angles. Discrete tomography is concerned with objects that consist of a small number of materials, which makes it possible to compute accurate reconstructions from highly limited projection data.  For cases where the allowed intensity values in the reconstruction are known a priori, the discrete algebraic reconstruction technique (DART) has shown to yield accurate reconstructions from few projections. However, a key limitation is that the benefit of DART diminishes as the number of different materials increases. Many tomographic imaging techniques can simultaneously record tomographic data at multiple \emph{channels}, each corresponding to a different weighting of the materials in the object. Whenever projection data from more than one channel is available, this additional information can potentially be exploited by the reconstruction algorithm. In this paper we present Multi-Channel DART (MC-DART), which deals effectively with multi-channel data. This class of algorithms is a generalization of DART to multiple channels and combines the information for each separate channel-reconstruction in a multi-channel segmentation step. We demonstrate that in a range of simulation experiments, MC-DART is capable of producing more accurate reconstructions compared to single-channel DART.
\end{abstract}
\section{Introduction} \label{sec:Intro}


Tomography is a non-invasive technique for creating 2D or 3D images of the inner structure of an object. Projections of the object are acquired by sending photonic or particle beams (e.g. X-rays, electrons, neutrons) through the object in a particular direction and measuring the signal resulting from interaction of the beam and the object at a detector. By acquiring this data from multiple positions and under various angles, a collection of projections is obtained. An image of the interior of the object is then reconstructed by applying a reconstruction algorithm to this projection data. Tomography is successfully used in many fields, including medical imaging \cite{Hsieh} and electron tomography in materials science \cite{Electronv2,Midgley}. If a large number of accurate projection images are available, solving the reconstruction problem is straightforward by a closed-form inversion formula \cite{Buzug}. Practical constraints on the dose, acquisition time or available space can impose limitations on the number of projections that can be taken, the angular range, or the noise level of the data, resulting in artefacts in the reconstructed images if standard reconstruction methods are used \cite{Hsieh}.

Discrete tomography is a powerful technique for dealing with such limited tomographic data. It can be applied if the object consists of only a limited number of materials with homogeneous densities. The Discrete Algebraic Reconstruction Technique (DART) \cite{DART1,DART2} is an algebraic reconstruction method for discrete tomography that alternates between continuous reconstruction steps and discretization of the image intensities by segmentation. The DART algorithm has demonstrated to obtain higher image quality reconstructions with limited projections and angles compared to standard reconstruction methods. Numerous successive studies have improved the DART algorithm, which include automatic parameter estimation (PDM-DART \cite{PDMDART} and TVR-DART \cite{TVRDART}), multi-resolution reconstruction (MDART \cite{MDART}), relaxing voxel constraints (SDART \cite{SDART}) and adaptive boundary reconstructions (ADART \cite{ADART}). Nevertheless, a key limitation of DART is that it can only improve reconstruction quality if the number of different materials in the object is relatively small. The main reason is that for a larger number of materials, the segmentation step is no longer effective \cite{DART1,DART2}.

\begin{figure}[!ht]
	\subcaptionbox{Material distribution \label{fig:1a}}[0.29\textwidth]{
		\includegraphics[width=0.29\textwidth]{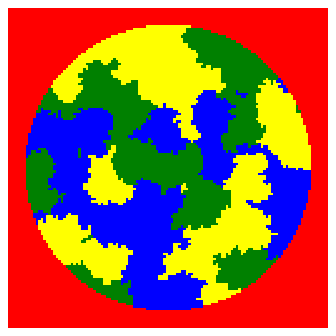}}
    \subcaptionbox{One-dimensional attenuations\label{fig:1b}}[0.32\textwidth]{
		\includegraphics[width=0.35\textwidth]{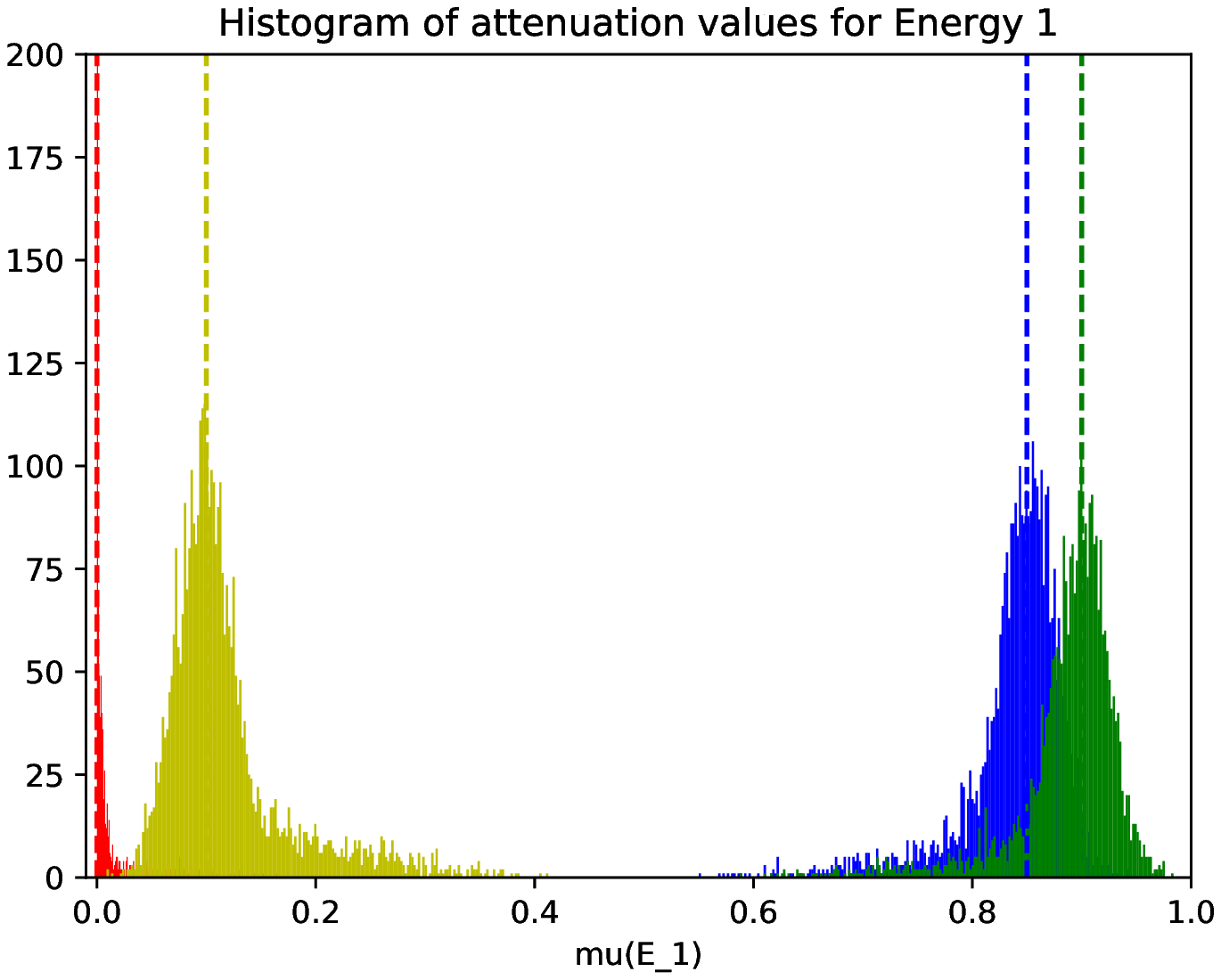}
		\includegraphics[width=0.35\textwidth]{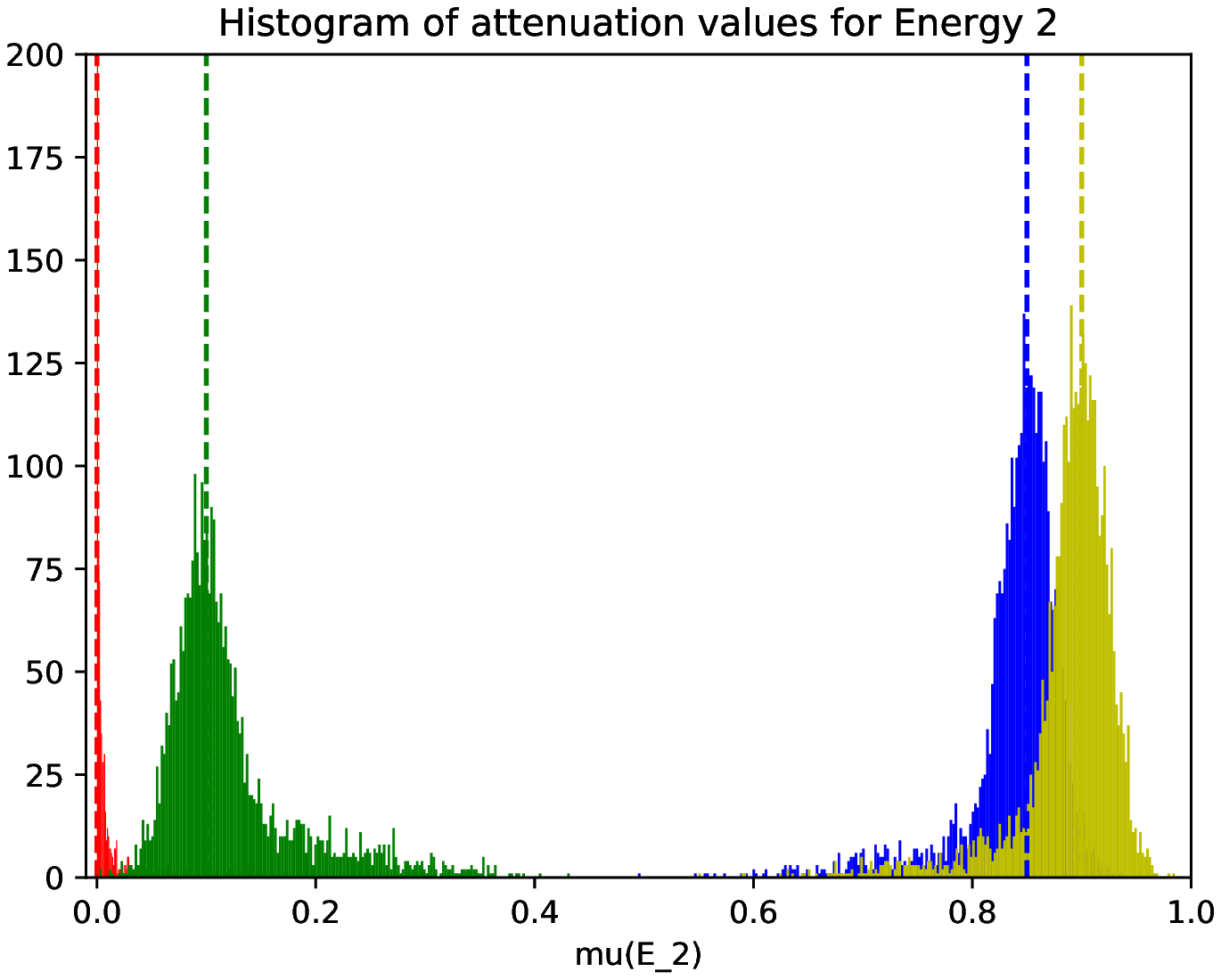}}
    \subcaptionbox{Two-dimensional attenuations \label{fig:1c}}[0.35\textwidth]{
		\includegraphics[width=0.35\textwidth]{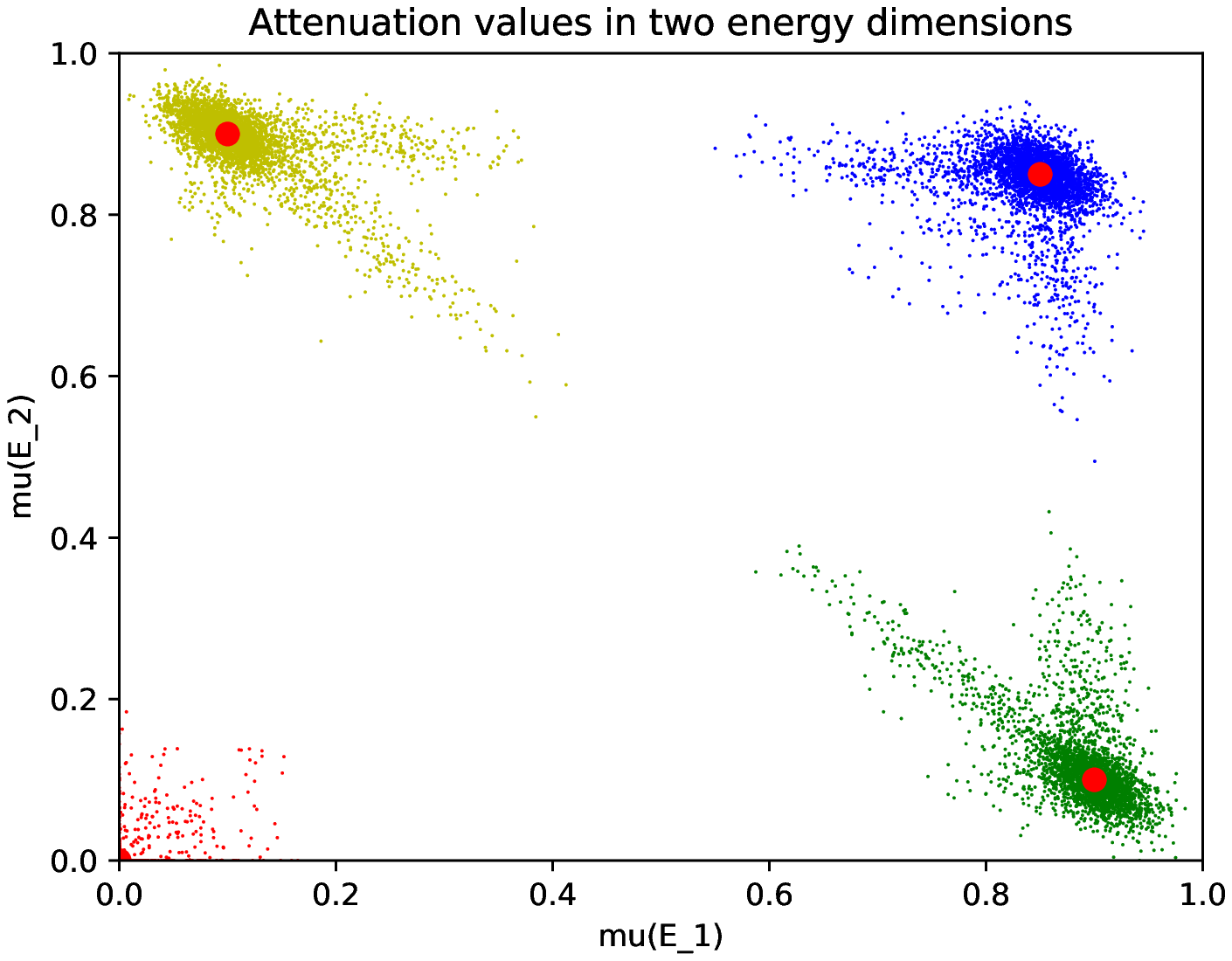}}
	\caption{Elementary example of separation difficulties during segmentation. (a) Distribution of the three materials (blue, yellow, green) in the object. The background is indicated in red. (b) Histogram of attenuation values of pixels at energy $E_1$ (above) and $E_2$ (below). Vertical lines show true material attenuations. (c) Attenuations of the materials (red dots) and computed attenuations by a reconstruction algorithm for each voxel (colors indicate the material they belong to).}\label{fig:1}
    \label{fig:Separation}
\end{figure}

In some cases it is possible to obtain tomographic information in multiple measurements channels. For instance, in X-ray imaging the beams are typically polychromatic, i.e. X-ray photon energies are distributed over a spectrum. Each material in the object has different attenuation properties for different X-ray energy levels. Whenever a single X-ray energy value is desired the range of energies within the beam can be narrowed by applying filters at the X-ray source \cite{Buzug}. Some detectors are capable of separating the incoming photons into energy bins while counting (e.g. HEXITEC \cite{HEXITEC}). In these cases spectral or multi-channel projection data is acquired, providing additional information about the object at different energies. Compared to the single-channel setting, where each material has a single attenuation value in the reconstructed image, in the multi-channel setting the attenuation value for each material varies along the channels. In this way, a tomographic dataset of the object is acquired for each channel, where the attenuation value of the materials changes throughout these datasets. This multi-channel imaging can potentially yield extra information about the materials. With more materials in the object, especially with similar attenuation features at a fixed energy, having data from multiple channels enables a better separation during segmentation. A conceptual example of this is shown in Figure~\ref{fig:Separation}. It is hard to separate points based on their attenuation values in a one-dimensional energy space. For instance, the right side of the blue area might as well be assigned to the green or yellow material during segmentation. With two energy dimensions the points are easily separable, since each voxel value lies close to its attenuation cloud center. Note that these spectra are artificial and not likely to occur in real-world examples.

In this paper we present a new class of algorithms that combines DART with multi-channel imaging for solving discrete multi-channel reconstruction problems. Our algorithm can combine the information from multiple channels to produce a final segmentation that is superior to that of the (single-channel) DART algorithm. Note that since this new method is designed by means of modules or subroutines that are interchangeable (as with DART), the method is essentially a class of algorithms providing a framework for dealing with multi-channel data. For simplicity, however, we will frequently call this framework an algorithm. 

This paper is structured as follows. Section 2 introduces the multi-channel discrete tomography problem. In Section 3 the DART algorithm is restated and the Multi-Channel DART (MC-DART) algorithm is introduced. Results of experiments with this algorithm are reported in Section 4. Finally, Section 5 presents the conclusions of this study.

\section{Problem Formulation} \label{sec:Problems}

The standard (single-channel) tomography problem can be modeled as a system of linear equations. The image is characterized by a vector of voxel attenuation values $\vec{x} \in \mathbb{R}^n$, where $n$ is the number of voxels. We will work with 2D images, but the problem formulation and methods in this paper can easily be extended to the 3D setting. We will refer to the image pixels as \emph{voxels} to distinguish these from detector pixels. We will often interchangeably speak of voxels and their corresponding indices. The projection values (also called data) are given as the vector $\vec{p} \in \mathbb{R}^l$, where $l$ is the number of projection angles times the number of detector pixels. The reconstruction problem can then be described by solving the following set of linear equations for $\vec{x}$:
\begin{align}
\vec{W}\vec{x} &= \vec{p} \label{Eq:OptProblem}.
\end{align}
Here $\vec{W}$ is the projection matrix, also called the forward operator~\cite{KakSlaney}. This matrix incorporates the contribution of each voxel to each projection, where element $w_{ij}$ indicates the contribution of voxel $j$ to projection $i$. Applying the operator $\vec{W}$ on a vector $\vec{x}$ results in the forward projection (also called sinogram). Since inverting the matrix $\vec{W}$ is computationally too expensive (or not even possible, for example when the problem is ill-posed) the reconstruction problem is to find a solution $\vec{x}^*$ whose forward projection $\vec{W}\vec{x}^*$ matches the projection data best with respect to some norm $||\cdot||$.
\begin{align}
\vec{x}^* = \argmin_{\vec{x} \in \mathbb{R}^n} || \vec{W}\vec{x} - \vec{p} || \label{Eq:InvProblem}
\end{align}
Since this is a least squares problem over $\mathbb{R}^n$, a solution always exists. For simplicity of notation we also assume that it is unique. A vector that encapsulates noise from real-world examples can also be modeled with \eqref{Eq:InvProblem}. In our experiments with phantom examples in Section~\ref{sec:Experiments} there is no noise.

In the \emph{discrete tomography problem}, the image to be reconstructed consists of a limited number of materials with homogeneous densities, each having an attenuation which is known beforehand by means of the set $\mathcal{R} = \{\rho_1, \ldots, \rho_m\}$, where $m$ is the number of different materials in the object. Therefore the problem to be solved becomes finding a vector $\vec{x} \in \mathcal{R}^n$ that matches the data best:
\begin{align}
\vec{x}^* = \argmin_{\vec{x} \in \mathcal{R}^n} || \vec{W}\vec{x} - \vec{p} ||. \label{Eq:DiscOptProblem}
\end{align}
Note that this is a minimization problem over a non-empty finite set. Hence, a minimum always exists. Again, it does not need to be unique but we use this notation throughout the paper for simplicity.

In the \emph{multi-channel} setting different properties of the target can be individually interrogated and measured. The information of each property is obtained through a separate channel. An example of channel is an energy level, as in the example in Section~\ref{sec:Intro}. In Figure~\ref{fig:1b}, the channels are the two energy levels revealing attenuations of the object at different energies. In a more abstract way the object is described as set a of voxels with labels instead of attenuation values, since each material has different attenuation values in different channels. The material labels are values in the set $\mathcal{M} = \{1, 2,\ldots, m\}$. The channel indices are given by $\mathcal{E} = \{E_1, E_2, \ldots, E_C\}$ where the number of channels is given by $C$. Again, the attenuations are known beforehand in the sets $\mathcal{R}_{E_1} = \{\rho_{1,1}, \ldots, \rho_{1,m}\}, \mathcal{R}_{E_2} = \{\rho_{2,1}, \ldots, \rho_{2,m}\}, \ldots, \mathcal{R}_{E_C} = \{\rho_{C,1} \ldots, \rho_{C,m}\}$. In this setting, let $\vec{\mathcal{R}} = \cup \mathcal{R}_{E_c}$. The function $\mu: \mathcal{M} \times \mathcal{E} \mapsto \vec{\mathcal{R}}$ maps the label-channel combinations to their attenuation value, so the attenuation of a material with label $s$ at channel $E_c$ is given by $\mu(s,E_c)$. Note that there is not necessarily a one-to-one correspondence between the attenuation values and the material-channel combinations, because some combinations can have the same attenuation value.
In this multi-channel case the projection data is given by a vector of projection data vectors at various channels:
\begin{align}
\vec{P} &= (\vec{p}_{E_1}, \ldots, \vec{p}_{E_C}) \in \mathbb{R}^{n\times C}. \label{Eq:ProjData}.
\end{align}
For each channel $E_c$ the reconstruction problem for $\vec{x}_{E_c}$ is given by the following set of linear equations:
\begin{align}
\vec{W}\vec{x}_{E_c} &= \vec{p}_{E_c}, && E_c \in \{E_1, \ldots E_C\}. \label{Eq:InvProblemMC}
\end{align}
For $\vec{y} \in \mathcal{M}^n$, define $\mu(\vec{y}, E_h) = (\mu(\vec{y}_1, E_h), \ldots, \mu(\vec{y}_n, E_h))^\top$ as the vector of voxel attenuation values at channel $E_h$. The multi-channel problem is now defined as follows. Given data vector $\vec{P}$ and projection matrix $\vec{W}$, find a labeling vector $\vec{y}^* \in \mathcal{M}^n$ such that for each channel $E_c$ the difference between forward projection $\vec{W}\mu(\vec{y}^*, E_h)$ and data is minimal with respect to some norm $||\cdot||$:

\begin{align}
\vec{y}^* = \argmin_{\vec{y} \in \mathcal{M}^n} \sum_{h=1}^C ||\vec{W}\mu(\vec{y}, E_h) - \vec{p}_{E_h} ||. \label{Eq:DiscOptProblemMC}
\end{align}
Note that for one channel the minimization problem is equivalent to \eqref{Eq:DiscOptProblem} where the labeling is given by the attenuation values $\vec{x}$, by setting $\mu(\vec{y}, E_1) = \vec{x}$ and $\mathcal{M}^n = \mathcal{R}^n$ and $\vec{p}_{E_1} = \vec{p}$: 

\begin{align}
\vec{y}^* &= \argmin_{\vec{y} \in \mathcal{M}^n} ||\vec{W}\mu(\vec{y}, E_1) - \vec{p}_{E_1} || \\
		  &= \argmin_{\vec{x} \in \mathcal{R}^n} ||\vec{W}\vec{x} - \vec{p} ||.
\end{align}

\section{Algorithms}

In this section the Multi-Channel DART (MC-DART) framework for solving the minimization problem of Eq.~\eqref{Eq:DiscOptProblemMC} is introduced. We first explain the DART algorithm as given in \cite{DART1} by discussing the overall structure and its building blocks. We then describe each building block of the MC-DART algorithm separately in more detail. Note that ASTRA~\cite{ASTRA2,ASTRA} provides an implementation for numerically computing all projection matrices in these algorithms, either by storing the full matrix or doing all necessary computation in a matrix-free way.

\subsection{DART}
\begin{figure}		
	\centering
	\includegraphics[width=1\textwidth]{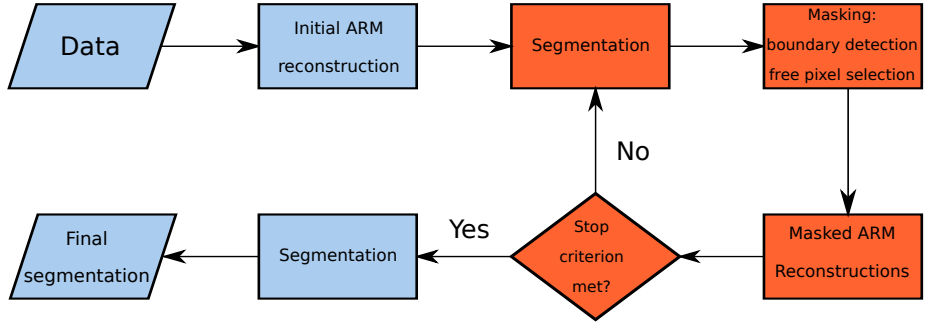}
	\caption{Flow chart of the DART algorithm. The DART iteration activities are indicated in red and the initialization and post-segmentation activities are indicated in blue.}\label{fig:FlowChartDART}		
\end{figure}

The DART algorithm attempts to solve the optimization problem of Eq. \eqref{Eq:DiscOptProblem} by iteratively alternating between continuous reconstruction steps and discrete segmentation steps. The number of materials in the object to be reconstructed and their attenuation values should be known beforehand, given by the function $\mu$. The algorithm consists of several phases, which are indicated in the flow-chart in Figure~\ref{fig:FlowChartDART}. The pseudocode of DART is given in Algorithm~\ref{Alg:Dart}.

\begin{algorithm}
\begin{algorithmic}[1]
\Statex \bf{Input}: $\vec{W}$, $\vec{p}$, $\mathcal{R}$
\State $\vec{x}^0 \gets\ \text{Mask-ARM}(\vec{W}, \vec{p}, \bf{1}_n, \bf{0}_n)$
\For {$k=1 \ \text{to} \ K$}
\State $\vec{y}^k \gets \ \text{Seg}(\vec{x}^k, \mathcal{R})$
\State $\vec{M}^k \gets \ \text{Mask}(\vec{y}^p)$
\State $\vec{x}^k \gets \ \text{Mask-ARM}(\vec{W}, \vec{p}, \vec{M}^k, \vec{x}^{k-1})$
\EndFor
\State $\textbf{Output:}\ \vec{x}^K, \text{Seg}(\vec{x}^K, \mathcal{R})$
\end{algorithmic}
\caption{DART}\label{Alg:Dart}
\end{algorithm}

\subsubsection{Initialization}
In the initialization phase, given the projection data $\vec{p}$ and the projection properties by means of $\vec{W}$, an initial reconstruction $\vec{x}^0$ is calculated using an Algebraic Reconstruction Method of choice (hereafter referred to as the ARM), for example ART, SART or SIRT \cite{KakSlaney}. With the initial reconstruction $\vec{x}^0$ at hand, the main loop of the DART algorithm begins. \\

\subsubsection{Segmentation}
In this main loop, in iteration $k$ the image $\vec{x}^{k-1}$ is segmented using a simple thresholding scheme, forming the image $\vec{y}^k \in \mathcal{R}^n$, by computing for every voxel $j$ the closest material attenuation value:
\begin{align}
\vec{y}_j^k	&= \begin{cases}
    			\rho_1, & \vec{x}_j^{k-1} < \frac{1}{2}(\rho_1 + \rho_2) \\
    			\rho_2, & \frac{1}{2}(\rho_1 + \rho_2) \leq \vec{x}_j^{k-1} < \frac{1}{2}(\rho_2 + \rho_3) \\
                \ \ \ \ \vdots \\
    			\rho_m, & \frac{1}{2}(\rho_{m-1} + \rho_{m}) \leq \vec{x}_j^{k-1}
  				\end{cases} \\
			&= \argmin_{\rho \in \mathcal{R}} ||\vec{x}^{k-1}_j - \rho||_2.
\end{align}
The second expression is easier to generalize to a higher-dimensional setting, which will we done in Section~\ref{sec:MCDART}.

\subsubsection{Boundary detection and masking}
A set of voxels in the figure is then selected for a new reconstruction to refine the resulting image. First, the set $B^k \subset \{1, \ldots, n\}$ of boundary voxel indices is determined based on the segmentation. Various schemes can be applied for boundary detection. Additionally, a set $U^k \subset \{1, \ldots, n\}$ of free voxel indices is determined, where each voxel is included with a certain probability $1-\beta$, with $0 \leq \beta \leq 1$. The process of selecting the voxels $U^k \cup B^k$ to be reconstructed and the voxels to be left out is called masking. Note that in the initialization phase all voxels are included in the mask.

\subsubsection{Masked ARM Reconstructions}
The set of free voxel indices $U^k \cup B^k$ are subjected to a new ARM reconstruction. This is done by computing the forward projection of the voxels $(\vec{y}_j^k)$ with $j \notin U^k \cup B^k$, and subtracting this from the input data $\vec{p}$ to obtain the residual sinogram $\overline{\vec{p}}^k$. The subproblem that has to be solved in this phase is:
\begin{align}
\overline{\vec{W}}^k\overline{\vec{x}}^k 	&= \overline{\vec{p}}^k \label{eq4}.
\end{align}
In Eq. \eqref{eq4} matrix $\overline{\vec{W}}^k$ is defined by $\overline{\vec{W}}^k = (w_{ij})_{ j \in U^k \cup B^k}$ and vector $\overline{\vec{x}}^k$ to be found has length $|U^k \cup B^k|$. Thus, the system of equations contains the same number of equations as Eq.~\eqref{Eq:InvProblem} but has fewer unknowns. The system is solved using a fixed number of ARM iterations, taking the values of $(\vec{x}^{k-1}_j)_{j \in U^k \cup B^k}$ as the starting condition. The complete reconstruction $\vec{x}^k$ at the end of iteration $k$ is then formed by merging $\overline{\vec{x}}^k$ with $\vec{y}^k$.

Some DART implementations also include a smoothing step at this point. The entire loop is repeated a predefined number of times. After the loop ends, the image is segmented one more time. Note that the DART algorithm has many degrees of freedom. This includes the number of ARM iterations in the initialization phase, the number of DART iterations, the number of ARM iterations during these DART iterations, the fixing probability $\beta$, and possibly parameters in the smoothing operation. The quality of the reconstructions also depends on the tomographic setup, such as the number of projections and the number of projection angles, and on the complexity of the object, including the number of materials and different attenuation values. Despite the DART algorithm performing well in practice, it is a heuristic method for which no solution guarantees exist \cite{Bounds}. The DART algorithm is also highly modular. Approaches for segmentation, boundary detection, reconstruction (ARM) and possible smoothing can easily be changed without sacrificing the overall structure of the algorithm. For the multi-channel algorithm proposed in this paper the segmentation phase is adapted to using all multi-channel reconstructions as input. 

The complexity of the framework depends on the algorithms that are used for reconstruction and segmentation. In this paper we use SIRT as the reconstruction algorithm and the thresholding segmentation as described above. Therefore, in this case, the DART algorithm has a time complexity of $O(Kn(m+l))$. The space complexity of our implementation is $O(ln)$.

\subsection{Multi-channel DART} \label{sec:MCDART}

\begin{figure}		
	\centering
	\includegraphics[width=1\textwidth]{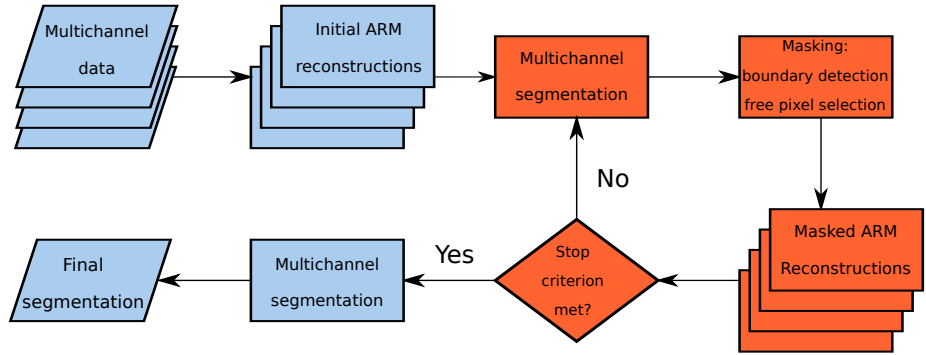}
	\caption{Flow chart of the MC-DART algorithm. A stacked number of activities indicate that they are applied at different channels simultaneously.}\label{fig:FlowChartMCDART}
\end{figure}

We now present the Multi-Channel DART (MC-DART) algorithm and outline its separate building blocks. Most focus will be on the multi-channel segmentation. Note that labeling single-channel images separately by attenuation values does not work here, since across multiple channels different materials can have the same attenuation. Therefore, there are some slight changes in the other blocks as well due to a new labeling mechanism. The algorithm structure is shown in the flow-chart in Figure~\ref{fig:FlowChartMCDART}. The pseudocode of MC-DART is given in Algorithm~\ref{Alg:MCDart}.

\begin{algorithm}
\begin{algorithmic}[1]
\Statex \bf{Input}: $\vec{W}$, $\mathcal{E}$, $\vec{P}$, $\vec{\mathcal{R}}$, $\mathcal{M}$, $\mu$
\For {$c=1 \ \text{to} \ C$}
\State $\vec{x}_{E_c}^0 \gets\ \text{Mask-ARM}(\vec{W}, \vec{p}_{E_c}, \bf{1}_n, \bf{0}_n)$
\EndFor
\For {$k=1 \ \text{to} \ K$}
\State $\vec{y}^k \gets \ \text{MCSeg}(\vec{X}^k, \vec{\mathcal{R}}, \mathcal{M}, \mu)$
\State $\vec{M}^k \gets \ \text{Mask}(\vec{y}^k)$
\For {$c=1 \ \text{to} \ C$}
\State $\vec{x}_{E_c}^k \gets \ \text{Mask-ARM}(\vec{W}, \vec{p}_{E_c}, \vec{M}^k, \vec{x}_{E_c}^{k-1})$
\EndFor
\EndFor
\State $\textbf{Output:}\ \vec{X}^K,\text{MCSeg}(\vec{X}^K, \vec{\mathcal{R}}, \mathcal{M}, \mu)$
\end{algorithmic}
\caption{MC-DART}\label{Alg:MCDart}
\end{algorithm}

\subsubsection{Initialization}
In the multi-channel setting we start out with a vector of projection data $\vec{P}$ at various channels and the matrix $\vec{W}$ as before. For each channel $E_c$ a reconstruction $x_{E_c}^0$ is computed using the selected ARM. This results in $C$ initial reconstructions for the MC-DART loop. 

\subsubsection{Multi-Channel Segmentation}
Given the reconstructions for all channels, similar to the DART segmentation, the multi-channel segmentation will determine a label image $\vec{y}^k \in \mathcal{M}^n$. Let $\vec{\mu}(s) = (\mu(s,E_1), \ldots, \mu(s,E_C)) \in \mathcal{R}_{E_1} \times \mathcal{R}_{E_2} \times \ldots \times \mathcal{R}_{E_C}$ be the vector of all attenuation values at each energy for material $s \in \mathcal{M}$, and let $X^k(\cdot,j) = (\vec{x}_{j,E_1}^k, \ldots, \vec{x}_{j,E_C}^k)$ be the vector of all attenuation values of voxel $j$ at each channel. We compute the segmented image by computing for each voxel $j \in \{1, \ldots, n\}$ the label using a basic thresholding scheme:
\begin{align}
\vec{y}_j^k	&=	\argmin_{s \in \mathcal{M}} ||X^k(\cdot, j) - \vec{\mu}(s)||_2 \label{eq8}.
\end{align}
Essentially, this operation selects the material label for which the multi-dimensional difference between the material attenuation and voxel attenuations is smallest.

\subsubsection{Masking and boundary detection}
The masking works exactly the same as in the single-channel case. Given the segmentation $\vec{y}^k$ the masking produces a set $U^k \cup B^k$ of voxel indices to be included in the multi-channel reconstructions.

\subsubsection{Multi-Channel Reconstructions}
In the MC-DART algorithm the reconstructions are handled separately for each energy. Thus, in MC-DART iteration $k$ the ARM is invoked $C$ times to find $\overline{\vec{x}}_{E_c}^k$ for each channel $c$ in
\begin{align}
\overline{\vec{W}}^k \overline{\vec{x}}_{E_c}^k 	&= \overline{\vec{p}}_{E_c}^k \label{eq7}.
\end{align}
The resulting (merged) reconstructions are then given by $\vec{X}^k := (\vec{x}_{E_1}^k, \ldots, \vec{x}_{E_C}^k) \in \mathbb{R}^{n \times C}$.

As with DART, the complexity of this framework depends on the reconstruction and segmentation methods that are chosen, as well as the extent of parallelization. If we use SIRT and the multi-channel segmentation method as described above and use a completely sequential implementation, the time complexity of MC-DART is $O(CKn(l+m))$. Because of the dependencies on the methods, we rather speak of a \emph{relative complexity} of MC-DART to DART, which we define as the ratio of the sequential MC-DART complexity to that of DART, irrespective of the subroutines used. This relative time complexity is $O(C)$. The space complexity of this algorithm instance of MC-DART is $O(Cn)$, resulting in a relative space complexity of $O(C)$ as well.

\section{Experimental Results} \label{sec:Experiments}
In this section the performance of the described MC-DART framework in terms of reconstruction and segmentation is presented. A series of experiments have been designed in which the number of channels $C$ and different materials $m$ are varied. For each experiment, multiple random phantoms are created. The size of these two-dimensional phantoms is $128 \times 128$ pixels, and each consists of a circular disk containing a random parcellation among $m$ materials in such a way that the total surface is approximately equal for each material. 
An example of this random phantom is given in Figure~\ref{fig:1a}, where $m=3$. Given the number of materials and channels, random attenuation spectra are generated by assigning a random number $\mu(s,E_c) \sim \mathcal{U}(0,1)$ for each channel-material combination, where $s \in \{1,\ldots,m\}$ and $E_c \in \{E_1,\ldots, E_C\}$. With this way of generating spectra no dependencies between channels are established. Note that in most practical applications such dependencies do exist, as materials all have their own attenuation spectrum.
For each phantom, reconstructions are made. The reference values for the tomographic setup and the parameter values of the MC-DART reconstruction algorithm for these reconstructions are summarized in Table~\ref{tab:Parameters}. For multi-channel segmentation the method as described in Section~\ref{sec:MCDART} is used.

\begin{table}[H]
\makebox[\textwidth][c]{
\begin{tabular}{|l|c|c|c|} \hline
\textbf{Parameter}		& \textbf{Reference value} 	\\ \hline
Angles					& 32 (equidistant)	\\
ARM 					& SIRT				\\
Start iterations		& 10   				\\                         
MC-DART iterations $K$	& 10				\\
ARM iterations			& 10				\\
Fix probability $\beta$ & 0.99				\\
\hline     
\end{tabular}}
\caption{Reference values for the parameters of the tomographic setup and the reconstructions algorithm for all experiments.}
\label{tab:Parameters}
\end{table}

We vary the number of channels $C \in\{1,\ldots,10\}$ and materials $m \in \{2,\ldots,10\}$ independently. For each combination, a random phantom $\vec{y}_{\text{init}}$ is created, after which data $\vec{P}$ is generated by applying the forward projection as described in Section~\ref{sec:Problems} on the phantom by applying $\mu$ and $\vec{W}$ on $\vec{y}_{\text{init}}$. In all experiments parallel-beam geometries are used and the detector size is $128$ pixels. After this, the MC-DART algorithm as described in Section~\ref{sec:MCDART} is applied with $K=10$ MC-DART iterations. The final segmentation is compared to the original phantom and the pixel error is computed, which is defined as the number of pixels in the final segmentation $\vec{y}^K$ that are labelled differently compared to the corresponding pixels in the original phantom $\vec{y}_{\text{init}}$. Only the pixels in the inner disk of the phantoms are taken into account. All experiments are repeated for and averaged over $100$ runs with different phantoms. 

The creation of random phantoms is implemented in \MATLAB. The remainder of the experiment setup scripts are implemented in Python. The reconstruction algorithms, including the MC-DART algorithm, are implemented in Python, where the ASTRA Toolbox \cite{ASTRA2,ASTRA} is used to take care of the ARM invocations and forward projections, including the masking in each MC-DART iteration and the creation of matrices $\vec{W}$ and $\overline{\vec{W}}^k$ based on the geometric properties.

\begin{figure}		
	\centering
	\includegraphics[width=0.7\textwidth]{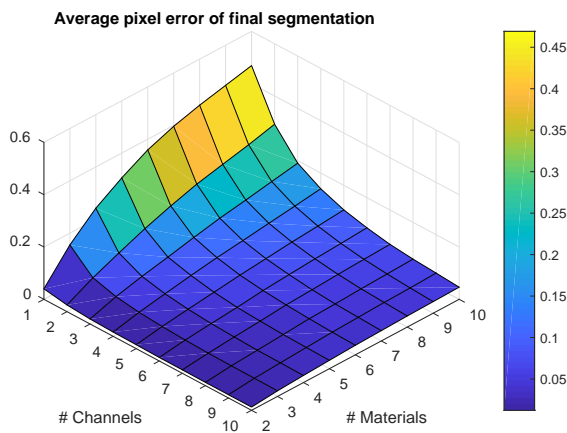}
	\caption{Pixel error percentage for different number of material-channel combinations.}\label{fig:StandardResults}
\end{figure}

\begin{figure}[!ht]	
	\centering
	\begin{subfigure}[!ht]{0.3\textwidth}
		\centering
		\includegraphics[width=\textwidth]{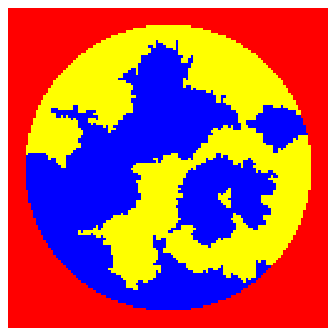}
		\caption{2-material phantom}\label{fig:Recsa}		
	\end{subfigure}
    \begin{subfigure}[!ht]{0.3\textwidth}
		\centering
		\includegraphics[width=\textwidth]{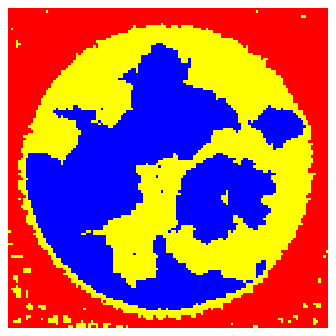}
		\caption{1-channel rec.}\label{fig:Recsc}
	\end{subfigure}
    \begin{subfigure}[!ht]{0.3\textwidth}
		\centering
		\includegraphics[width=\textwidth]{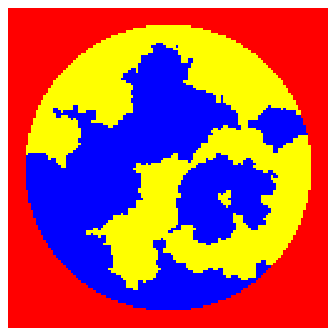}
		\caption{10-channel rec.}\label{fig:Recse}
	\end{subfigure}
    \begin{subfigure}[!ht]{0.3\textwidth}
		\centering
		\includegraphics[width=\textwidth]{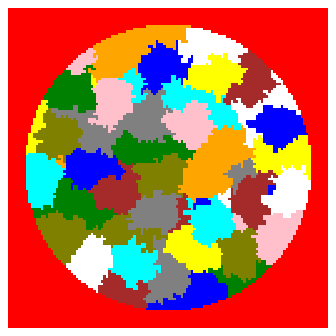}
		\caption{10-material phantom}\label{fig:Recsb}
	\end{subfigure}
    \begin{subfigure}[!ht]{0.3\textwidth}
		\centering
		\includegraphics[width=\textwidth]{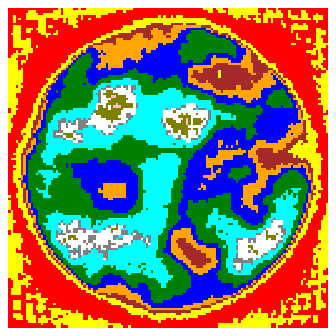}
		\caption{1-channel rec.}\label{fig:Recsd}
	\end{subfigure}
    \begin{subfigure}[!ht]{0.3\textwidth}
		\centering
		\includegraphics[width=\textwidth]{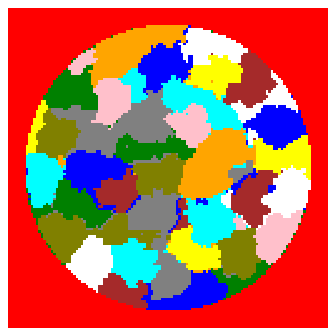}
		\caption{10-channel rec.}\label{fig:Recsf}
	\end{subfigure}
    \caption{Reconstructions for various setups. (a, d) Phantoms used with two and ten materials respectively. (b, e) Reconstructions using one channel. The mislabeled yellow pixels are because the attenuation of the yellow material is very close to zero. (c, f) Reconstructions using ten channels.}\label{fig:Recs}
\end{figure}

Figure~\ref{fig:StandardResults} shows the percentage of misclassified pixels with respect to the number of pixels in the inner disk. The percentage is lowest when the number of materials is low and the number of channels is high, while the percentage is highest when the number of channels is low and the number of materials is high. Given a number of channels, the percentage seems to scale logarithmically with the number of materials. On the other hand, given a number of materials, the percentage seems to scale exponentially with the number of channels for larger number of materials. Therefore, in this setup, the addition of only a few channels improves the reconstruction quality considerably. Figure~\ref{fig:Recs} shows examples of the reconstructions at the corners of the curved plane of Figure~\ref{fig:StandardResults}.

We have investigated the effect of changing the parameters that are shown in Table~\ref{tab:Parameters}. The number of starting iterations has no effect on the pixel error percentage curve. For these parameters, we found that increasing the number of MC-DART iterations further than $4$ had no significant effect on the reconstructions. This threshold depends on the number of ARM iterations in each MC-DART iteration. Also, the quality of the reconstructions increases only marginally when $\beta$ is increased. However, the pixel error percentage drops considerably as the number of ARM iterations during an MC-DART iteration increases. Also, when scanning data from many angles is available the reconstruction quality improvement with multiple materials become much better. For only $2$ angles, the reconstruction between $C=1$ and $C=10$ channels improves from pixel error percentage $27\%$ to $23\%$ for two materials and from $55\%$ to $41\%$ for ten materials. In comparison, for as much as $128$ angles the reconstructions between $C=1$ and $C=10$ channels improve by from $3\%$ to less than $1\%$ for two materials and from $46\%$ to $4\%$ for ten materials. We conclude that in all these cases the MC-DART algorithm gives better results when more channels are available.\\

\begin{figure}
	\centering
	\begin{subfigure}[!ht]{0.4\textwidth}
		\centering
		\includegraphics[width=\textwidth]{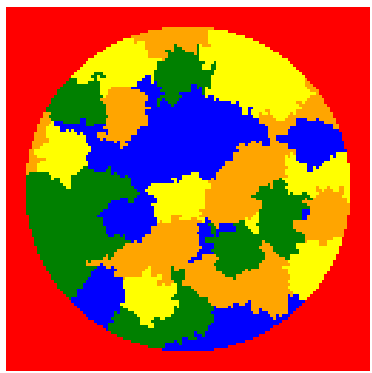}
		\caption{Phantom}\label{fig:4a}		
	\end{subfigure}
	\quad
	\begin{subfigure}[!ht]{0.48\textwidth}
		\centering
		\includegraphics[width=\textwidth]{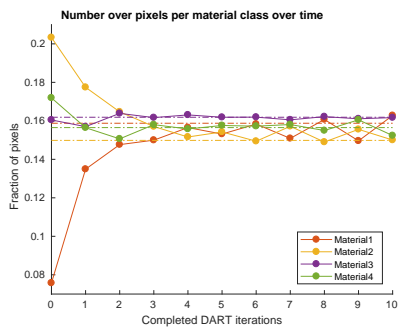}
		\caption{Number of pixels per class for DART}\label{fig:4b}
	\end{subfigure}
    
    \begin{subfigure}[!ht]{0.48\textwidth}
		\centering
		\includegraphics[width=\textwidth]{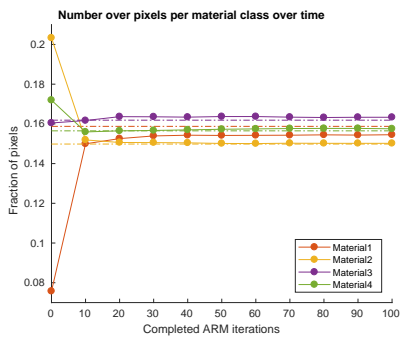}
		\caption{Number of pixels per class for non-DART}\label{fig:4c}
	\end{subfigure}
    \begin{subfigure}[!ht]{0.48\textwidth}
		\centering
		\includegraphics[width=\textwidth]{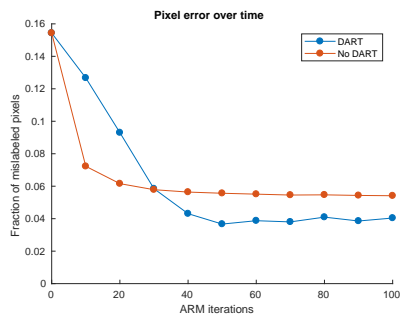}
		\caption{Pixel error over time}\label{fig:4d}
	\end{subfigure}
	\caption{Convergence behavior for material classes with $C=10$ channels. (a) Phantom that has been used with $m=4$ materials (b) Graph showing the behavior for each material class in a DART routine for this phantom. The number of iterations for the initial reconstruction is set to $2$, the number of DART iterations is $10$, the number of ARM iterations is set to $10$, the number of angles is $8$, fixing probability is set to $\beta=0.99$. The chosen ARM is SIRT. Shown are the number of pixels assigned per class during segmentation after each DART step, with the true value of these indicated by dashed lines. (c) Number of pixel assigned per class over number of ARM iterations. The number of DART iterations is 0, and instead we apply $100$ ARM iterations with $2$ initial iterations. The results are based on intermediate segmentations after each 10 ARM iterations, but these segmentations are not used in further iterations. Background pixels are excluded from the results (d) Pixel error over number of ARM iterations for both approaches.}\label{fig:4}
    \label{fig:EXP2}
\end{figure}

Additionally, apart from the pixel error, we investigate how the number of assigned pixels per material class behave as the MC-DART reconstruction proceeds. The results are shown in Figure~\ref{fig:4}. A random phantom with four different materials and background is used. The number of channels is set to $C=10$, and for each channel $c$ and material $m$ a random attenuation value $\mu(s,E_c) \sim \mathcal{U}(0,1)$ is generated. Then the MC-DART algorithm is applied to this phantom in two different experimental setups. In the first experiment, the number of MC-DART iterations is set to $10$ and the number of ARM iterations per MC-DART iteration is set to $10$. After each MC-DART iteration, the number of pixels assigned are calculated for each class. During the first four MC-DART iterations the number of assigned pixels is converging towards their real values. After this, the graphs enter an oscillatory phase in which for each class the number of assigned pixels alternates between two values whose average is not necessary the real number of pixels for that class. For comparison, in the second experiment the same setup is used, but without using MC-DART iterations and applying the same ARM for $100$ iterations instead. In this way the ARM is effectively invoked equally often. After each $10$ iterations a segmentation is made based on the current reconstruction and the pixels per class are measured, but no new forward projections are calculated from these segmentations and used in subsequent ARM iterations. In this case the number of pixels converges much more quickly for each class. Also, there is no oscillatory phase and the number of pixels are just as close to their true values as with the DART approach. However, plotting the total pixel error over time reveals that the pixel error in the non-DART case is higher. The pixel error for the MC-DART case needs more time to stabilize to its oscillatory phase, but the values are eventually lower than in the non-MC-DART case.

\section{Conclusions}
A new class of algorithms for solving discrete multi-channel reconstruction problems has been proposed. This framework uses the strength of DART regarding dealing with limited data in a multi-channel setting by using a multi-channel segmentation method. The experiments have shown that combining information from different channels by a multi-channel segmentation method increases the reconstruction quality compared to the single-channel DART algorithm. Therefore, we conclude that the MC-DART framework is a promising approach for dealing with multi-channel data.

\section{Discussion}

This paper presents the first steps to implement a multi-channel reconstruction technique using multi-channel segmentation. Currently, there are no standard approaches for the discrete multi-channel problem presented in Section~\ref{sec:Problems}. We propose a framework in which reconstruction and segmentation techniques can be exchanged. The modules in the framework can be adjusted to the problem to be solved. For instance, segmentation can be performed with neural network-based methods. The proposed method is not aimed at optimizing reconstructions with state-of-the-art ARMs or segmentation techniques but at presenting a framework to work with multi-channel data. If more data from different channels is available, this implementation outperforms DART but it does not mean that the problem is optimally solved.
To further develop this technique and transfer it to real-world settings, real-data properties should be taken into account. These properties include the correlation of attenuation values between channels and noise contained in the projection data.
In our study we only make use of the multi-channel data during segmentation. Another approach could be to use the multi-channel data during reconstruction, modeling the reconstruction problem as a large inverse problem where the unknowns are the material concentrations in each pixel (e.g. see \cite{kazantsev2018joint,tairi2016simultaneous}). However, solving this problem is much more involved and the MC-DART framework presented in this paper provides a simple but effective alternative of separating materials using multi-channel data.

\bibliographystyle{splncs03.bst}
\bibliography{biblio}

\begin{thebibliography}{10}
\providecommand{\url}[1]{\texttt{#1}}
\providecommand{\urlprefix}{URL }

\bibitem{PDMDART}
van Aarle, W., Batenburg, K.J., Sijbers, J.: Automatic parameter estimation for
  the discrete algebraic reconstruction technique (dart). IEEE Transactions on
  Image Processing  21(11),  4608--4621 (2012)

\bibitem{ASTRA2}
van Aarle, W., Palenstijn, W.J., Cant, J., Janssens, E., Bleichrodt, F.,
  Dabravolski, A., De~Beenhouwer, J., Batenburg, K.J., Sijbers, J.: Fast and
  flexible x-ray tomography using the astra toolbox. Optics express  24(22),
  25129--25147 (2016)

\bibitem{Bounds}
Batenburg, K.J., Fortes, W., Hajdu, L., Tijdeman, R.: Bounds on the quality of
  reconstructed images in binary tomography. Discrete Applied Mathematics
  161(15),  2236--2251 (2013)

\bibitem{DART1}
Batenburg, K.J., Sijbers, J.: Dart: a fast heuristic algebraic reconstruction
  algorithm for discrete tomography. In: Image Processing, 2007. ICIP 2007.
  IEEE International Conference on. vol.~4, pp. IV--133. IEEE (2007)

\bibitem{DART2}
Batenburg, K.J., Sijbers, J.: Dart: a practical reconstruction algorithm for
  discrete tomography. IEEE Transactions on Image Processing  20(9),
  2542--2553 (2011)

\bibitem{SDART}
Bleichrodt, F., Tabak, F., Batenburg, K.J.: Sdart: An algorithm for discrete
  tomography from noisy projections. Computer Vision and Image Understanding
  129,  63--74 (2014)

\bibitem{Buzug}
Buzug, T.M.: Computed tomography: from photon statistics to modern cone-beam
  CT. Springer Science \& Business Media (2008)

\bibitem{MDART}
Dabravolski, A., Batenburg, K.J., Sijbers, J.: A multiresolution approach to
  discrete tomography using dart. PloS one  9(9),  e106090 (2014)

\bibitem{Electronv2}
Frank, J.: Electron tomography. Springer (1992)

\bibitem{Hsieh}
Hsieh, J., et~al.: Computed tomography: principles, design, artifacts, and
  recent advances. SPIE Bellingham, WA (2009)

\bibitem{KakSlaney}
Kak, A.C., Slaney, M., Wang, G.: Principles of computerized tomographic
  imaging. Medical Physics  29(1),  107--107 (2002)

\bibitem{kazantsev2018joint}
Kazantsev, D., J{\o}rgensen, J.S., Andersen, M.S., Lionheart, W.R., Lee, P.D.,
  Withers, P.J.: Joint image reconstruction method with correlative
  multi-channel prior for x-ray spectral computed tomography. Inverse Problems
  34(6),  064001 (2018)

\bibitem{ADART}
Maestre-Deusto, F.J., Scavello, G., Pizarro, J., Galindo, P.L.: Adart: An
  adaptive algebraic reconstruction algorithm for discrete tomography. IEEE
  Transactions on image processing  20(8),  2146--2152 (2011)

\bibitem{Midgley}
Midgley, P., Weyland, M.: 3d electron microscopy in the physical sciences: the
  development of z-contrast and eftem tomography. Ultramicroscopy  96(3-4),
  413--431 (2003)

\bibitem{ASTRA}
Palenstijn, W.J., Batenburg, K.J., Sijbers, J.: The astra tomography toolbox.
  In: 13th International Conference on Computational and Mathematical Methods
  in Science and Engineering, CMMSE. vol. 2013, pp. 1139--1145 (2013)

\bibitem{tairi2016simultaneous}
Tairi, S., Anthoine, S., Morel, C., Boursier, Y.: Simultaneous reconstruction
  and separation in a spectral ct framework. In: Nuclear Science Symposium,
  Medical Imaging Conference and Room-Temperature Semiconductor Detector
  Workshop (NSS/MIC/RTSD), 2016. pp. 1--4. IEEE (2016)

\bibitem{HEXITEC}
Wilson, M., Dummott, L., Duarte, D., Green, F., Pani, S., Schneider, A.,
  Scuffham, J., Seller, P., Veale, M.: A 10 cm$\times$ 10 cm cdte spectroscopic
  imaging detector based on the hexitec asic. Journal of Instrumentation
  10(10),  P10011 (2015)

\bibitem{TVRDART}
Zhuge, X., Palenstijn, W.J., Batenburg, K.J.: Tvr-dart: a more robust algorithm
  for discrete tomography from limited projection data with automated gray
  value estimation. IEEE Transactions on Image Processing  25(1),  455--468
  (2016)

\end{thebibliography}

\end{document}